\documentclass[10pt,twocolumn,letterpaper]{article}

\makeatletter
\@namedef{ver@everyshi.sty}{}
\makeatother

\usepackage[pagenumbers]{cvpr}              

\usepackage{graphicx}
\usepackage{amsmath}
\usepackage{amssymb}
\usepackage{booktabs}
\usepackage{multirow}

\usepackage{amssymb}
\usepackage{bbding}
\usepackage{pifont}
\usepackage{wasysym}
\usepackage{utfsym}

\usepackage[pagebackref,breaklinks,colorlinks]{hyperref}

\usepackage[capitalize]{cleveref}
\crefname{section}{Sec.}{Secs.}
\Crefname{section}{Section}{Sections}
\Crefname{table}{Table}{Tables}
\crefname{table}{Tab.}{Tabs.}
\usepackage{bm}

\def\cvprPaperID{3692} 
\def\confName{CVPR}
\def\confYear{2023}

\begin{document}

\title{Prototype-based Embedding Network for Scene Graph Generation}

\author{
Chaofan Zheng\thanks{Equal contribution.}
\and Xinyu Lyu$^{\ast}$ 
\and Lianli Gao\thanks{Corresponding author.} 
\and Bo Dai
\and Jingkuan Song \\
 School of Computer Science and Engineering, \\University of Electronic Science and Technology of China, China\\
 {\tt\small \href{mailto:zheng_chaofan@foxmail.com}{zheng\_chaofan@foxmail.com},
\href{mailto:lianli.gao@uestc.edu.cn}{lianli.gao@uestc.edu.cn}
}
}


\maketitle
\begin{abstract}

Current Scene Graph Generation (SGG) methods explore contextual information to predict relationships among entity pairs.
However, due to the diverse visual appearance of numerous possible subject-object combinations, 
there is a large \textbf{intra-class variation} within each predicate category, \eg, ``man-\underline{eating}-pizza, giraffe-\underline{eating}-leaf'', 
and the severe \textbf{inter-class similarity} between different classes, \eg, ``man-\underline{holding}-plate, man-\underline{eating}-pizza'', in model's latent space. 
The above challenges prevent current SGG methods from acquiring robust features for reliable relation prediction.
In this paper, we claim that the predicate's category-inherent semantics can serve as class-wise prototypes in the semantic space for relieving the challenges.
To the end, we propose the \textbf{Prototype-based Embedding Network (PE-Net)}, which models entities/predicates with prototype-aligned compact and distinctive representations and thereby establishes matching between entity pairs and predicates in a common embedding space for relation recognition.
Moreover, \textbf{Prototype-guided Learning (PL)} is introduced to help PE-Net efficiently learn such entity-predicate matching, and \textbf{Prototype Regularization (PR)} is devised to relieve the ambiguous entity-predicate matching caused by the predicate's semantic overlap.
Extensive experiments demonstrate that our method gains superior relation recognition capability on SGG, achieving new state-of-the-art performances on both Visual Genome and Open Images datasets. The codes are available at
\url{https://github.com/VL-Group/PENET}.


\end{abstract}

\vspace{-1em}
\section{Introduction}
\label{sec:intro}

Scene Graph Generation (SGG) is a fundamental computer vision task that involves detecting the entities and predicting their relationships in an image to generate a scene graph, where nodes indicate entities and edges indicate relationships between entity pairs. Such a graph-structured representation is helpful for downstream tasks such as Visual Question Answering~\cite{vqa,vqa1,cfm:vqa1}, Image Captioning~\cite{cap1,cap2,cfm:cap1, cfm:videocap1}, and Image Retrieval~\cite{imgtrv1,imgtrv2,cfm:imgrtv1}. 

\begin{figure}[t]
  \center
  \includegraphics[width=1\linewidth]{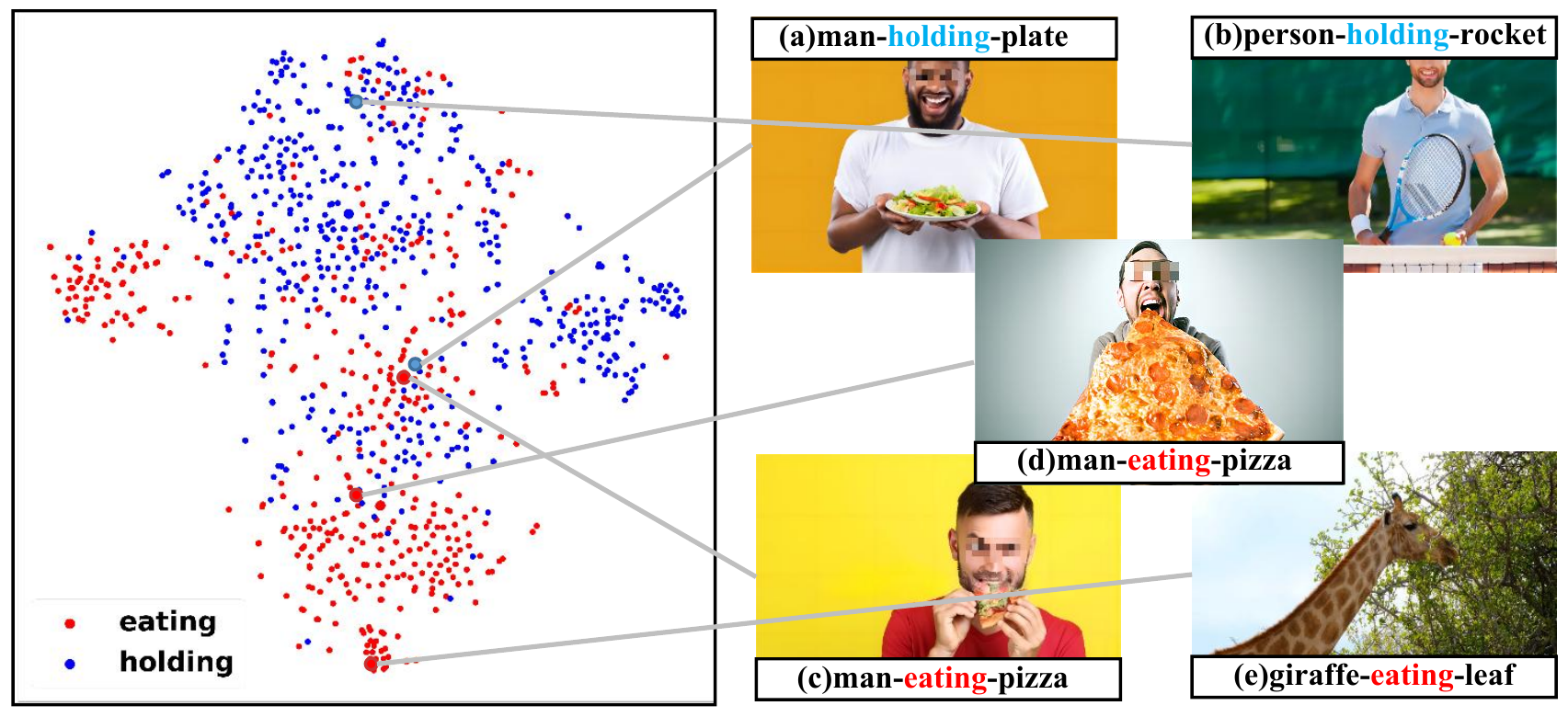}
\caption{The illustration of relation representations with large intra-class variation and severe inter-class similarity.  \textbf{Left:} the feature distribution of ``eating'' (in red)  and ``holding'' (in blue) obtained by Motifs~\cite{sgg:motifs}. \textbf{Right:} some instances of ``eating" and ``holding". 
Examples (c) and (e) illustrate that relation instances from the same class have diverse appearance.
Moreover, examples (a) and (c) demonstrate that similar-looking relation instances may belong to different categories.
}
\vspace{-1em}
\label{fig:intro}
\end{figure}

Existing SGG models~\cite{sgg:os-sgg,sgg:motifs,sgg:vctree,sgg:bgnn,sgg:hlnet,sgg:gcl,sgg:kern} typically start with an object detector that generates a set of entity proposals and corresponding features. Then, entity features are enhanced by exploring the contextual information taking advantage of message-passing modules. Finally, these refined entity features are used to predict pairwise relations. 
Although many works have made great efforts to explore the contextual information for robust relation recognition, they still suffer from biased-prediction problems, preferring common predicates (\eg,``on'', ``of'') instead of fine-grained ones (\eg,``walking on'', ``covering'').
To address the problem, various de-biasing frameworks~\cite{sgg:fgpl,sgg:cogtree,sgg:tde,sgg:ba-sgg,sgg:pcpl,sgg:ppdl,sgg:sgg-ht,sgg:dhl} have been proposed to obtain balanced prediction results. 
While alleviating the long-tailed issue to some extent, most of them only achieve a trade-off between head and tail predicates.
In other words, they sacrifice the robust representations learned on head predicates for unworthy improvements in the tail ones~\cite{sgg:dhl}, which do not truly improve the model's holistic recognition ability for most of the relations.  



The origin of the issue lies in the fact that current SGG methods fail to capture compact and distinctive representations for relations. 
For instance, as shown in Fig.~\ref{fig:intro}, the relation representation, derived from Motifs' latent space, is heavily discrete and intersecting. 
Hence, it makes existing SGG models hard to learn perfect decision boundaries for accurate predicate recognition. 
Accordingly, we summarize the issue as two challenges: large \textbf{Intra-class variation} within the same relation class and severe \textbf{Inter-class similarity} between different categories.

\noindent\textbf{Intra-class variation.} The intra-class variation arises from the diverse appearance of entities and various subject-object combinations.
Specifically, entities' visual appearances change greatly even though they belong to the same class.
Thus, represented as the union feature containing subject and object entities, relation representations significantly vary with the appearances of entity instances, \eg, various visual representations for ``pizza'' in Fig.~\ref{fig:intro}(c) \textit{vs.} Fig.~\ref{fig:intro}(d).
Besides, the numerous subject-object combinations of predicate instances further increase the variation within each predicate class, \eg, ``man-eating-pizza'' \textit{vs.} ``giraffe-eating-leaf'' in Fig.~\ref{fig:intro}(c) and Fig.~\ref{fig:intro}(e). 
%


\noindent\textbf{Inter-class similarity.} The inter-class similarity of relations originates from similar-looking interactions but belongs to different predicate classes. For instance, as shown in Fig.~\ref{fig:intro}(a) and Fig.~\ref{fig:intro}(c), the similar visual appearance of interactions between ``man-pizza'' and ``man-plate'' make current SGG models hard to distinguish ``eating'' from ``holding'', even if they are semantic irrelevant to each other.

The above challenges motivate us to study two problems:
1) For the intra-class variation, how to capture category-inherent features, producing compact representations for entity/predicate instances from the same category.
%
Moreover, 2) for the inter-class similarity, how to derive distinctive representations for effectively distinguishing similar-looking relation instances between different classes.  
%
%
Our key intuition is that semantics is more reliable than visual appearance when modeling entities/predicates.
%
Intuitively, although entities/predicates of the same class significantly vary in visual appearance, they all share the representative semantics, which can be easily captured from their class labels.
Dominated by the representative semantics, the representations of entities and predicates have smaller variations within their classes in the semantic space.
Besides, the class-inherent semantics is discriminative enough for visual-similar instances between different categories. 
Therefore, in conjunction with the above analysis, modeling entities and predicates in the semantic space can provide highly compact and distinguishable representations against intra-class variation and inter-class similarity challenges.


Inspired by that, we propose a simple but effective method, Prototype-based Embedding Network (PE-Net), which produces compact and distinctive entity/predicate representations for relation recognition. 
To achieve that, the PE-Net models entity and predicate instances with compact and distinguishable representations in the semantic space, which are closely aligned to their semantic prototypes.
Practically, the prototype is defined as the representative embedding for a group of instances from the same entity/predicate class. 
%
Then, the PE-Net establishes matching between entity pairs (\ie, subject-object ($\bm s,\bm o$)) and their corresponding predicates ($\bm p$) for relation recognition 
(\ie, $\mathcal{F}(\bm{s}, \bm{o}) \approx \bm{p}$).
Besides, a Prototype-guided Learning strategy (PL) is proposed to help PE-Net efficiently learn this entity-predicate matching. 
Additionally, to alleviate the ambiguous entity-predicate matching caused by the semantic overlap between predicates (\eg, ``walking on'' and ``standing on''), Prototype Regularization (PR) is proposed to encourage inter-class separation between predicate prototypes for precise entity-predicate matching.
%
Finally, we introduce two metrics, \ie, Intra-class Variance (IV) and Intra-class to Inter-class Variance Ratio (IIVR), to measure the compactness and distinctiveness of entity/predicate representations, respectively.


In summary, the main contributions of our work are three folds:
\begin{itemize}
    \item We propose a simple yet effective method, \ie, Prototype-based Embedding Network (PE-Net), which produces compact and distinctive entity/predicate representations and then establishes matching between entity pairs and predicates for relation recognition.
    \item Moreover, Prototype-guided Learning (PL) is introduced to help PE-Net efficiently learn such entity-predicate matching, and Prototype Regularization (PR) is devised to relieve the ambiguous entity-predicate matching caused by the predicate's semantic overlap.
    \item Evaluated on the Visual Genome and Open Images datasets, our method significantly increases the relation recognition ability for SGG, achieving new state-of-the-art performances.
\end{itemize}

\section{Related Work}
\label{sec:related_work}
We categorize the related works of SGG into the following fields: Vanilla Scene Graph Generation Model and Unbiased Scene Graph Generation Framework. 

\noindent \textbf{Vanilla Scene Graph Generation Model}. Numerous models have been proposed to solve the scene graph generation task from different perspectives in recent years. Early methods~\cite{sgg:vrd} attempted to detect objects and relations with independent networks, ignoring the rich contextual information. Afterward, ~\cite{sgg:imp} firstly proves that the contextual information can significantly improve the relation prediction and hence introduces an iterative message-passing mechanism to refine the features of objects and relations. ~\cite{sgg:motifs} further emphasizes the importance of contextual information between objects and utilizes the BiLSTM to encode the object and edge contextual information. Moreover, to avoid suffering from noisy information during message passing,~\cite{sgg:vctree} and~\cite{sgg:graphrcnn} design sparse structures to improve the model's context modeling capability. 
In addition, prior knowledge is also helpful for relation prediction. ~\cite{sgg:motifs} explores the statistical patterns of object co-occurrence for refining relation predictions. Besides, ~\cite{sgg:os-sgg} encodes the commonsense knowledge into the model to improve the few-shot recognition ability. 
However, due to the imbalanced data distribution, the vanilla SGG models struggle to recognize the fine-grained tail predicates.

\noindent\textbf{Unbiased Scene Graph Generation Framework.}
Recently, various de-biasing SGG frameworks have been proposed to tackle the biased predictions problem. ~\cite{sgg:tde} proposes a counterfactual causality method to remove the effect of context bias. ~\cite{sgg:cogtree} constructs a hierarchical tree structure from the cognitive perspective to make the tail predicates receive more attention. ~\cite{sgg:bgnn} compensates the disadvantages of over-sampling and under-sampling and proposes a bi-level sampling method. ~\cite{sgg:ba-sgg} creates a balanced learning process by constructing a balanced predicate learning space and semantic adjustment. ~\cite{sgg:nice} explicitly cleans the noisy annotations on the datasets to balance the data distribution. ~\cite{sgg:fgpl} introduces a predicate lattice to figure out the fine-grained predicate pairs that are hard to distinguish. Despite alleviating the biased problem to some extent, these methods improve the prediction performance of tail predicates at the expense of head ones, which do not truly improve the model’s holistic recognition ability.

Our work generates compact and distinctive entity/predicate representations by utilizing a prototype-based modeling method and cleverly-designed learning strategies, which achieves superior relation recognition performance on both head predicates and tail ones with a simple but effective framework. 



\begin{figure}[t]
  \center
  \includegraphics[width=1\linewidth]{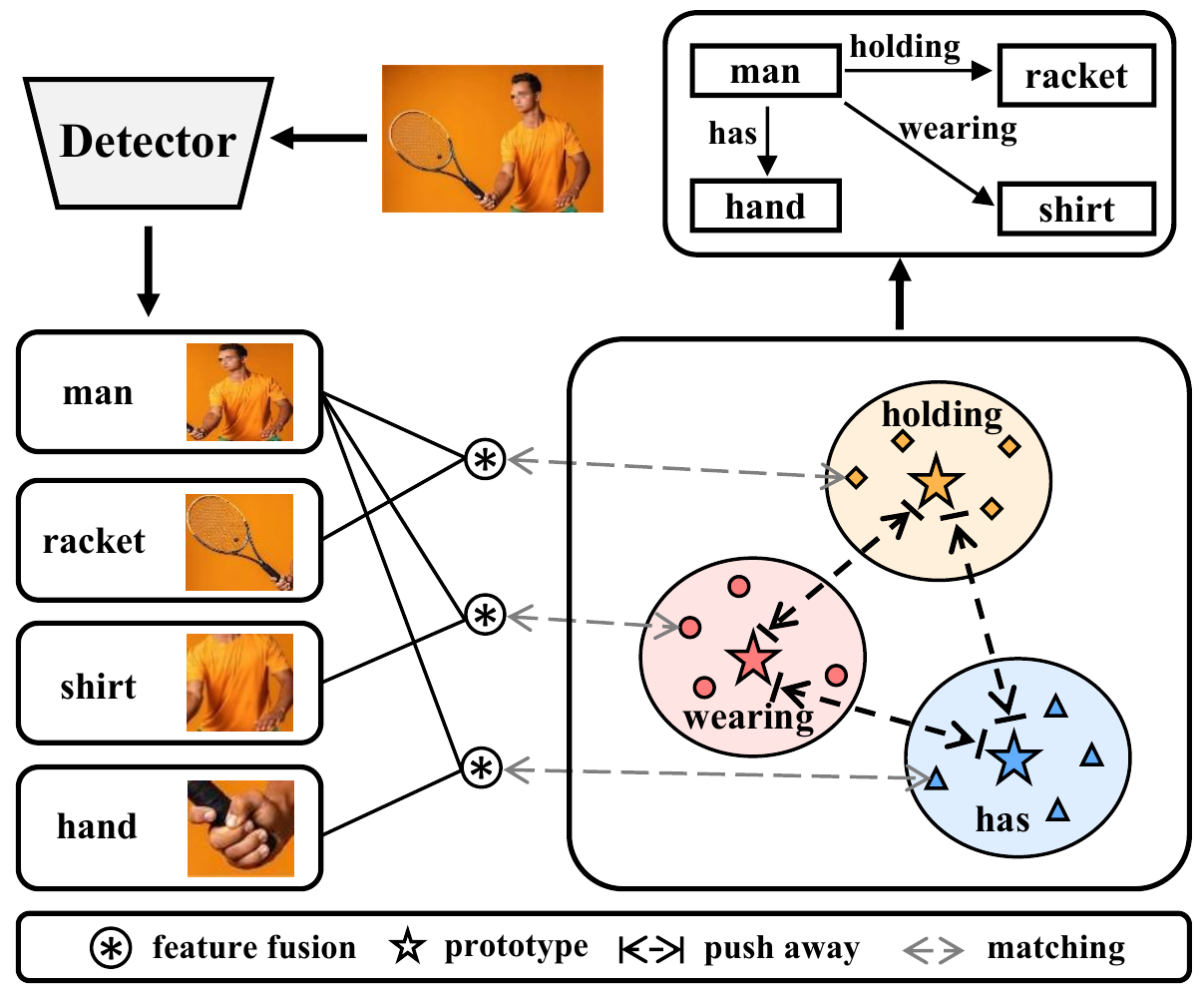}
\caption{The main process of our proposed PE-Net from the input image to the scene graph.}
\vspace{-1em}
\label{fig:framework}
\end{figure}





\section{Method}
\label{sec:method}

The whole pipeline of our Prototype-based Relation Embedding (PE-Net) is illustrated in~\cref{fig:framework}.
Following the previous works~\cite{sgg:sggbenchmark,sgg:tde}, we utilize an object detector (\eg, Faster R-CNN~\cite{fasterrcnn}) to generate a set of entity proposals with corresponding features.
Moreover, the features extracted from the union box between two entities are used to represent their corresponding predicates.
Given entity and predicate features, the Prototype-based Embedding Network (PE-Net)
models subject ($\bm s$), object ($\bm o$), and predicate ($\bm p$) instances with prototype-based compact and distinguishable representations.
Then, the PE-Net matches subject-object pairs ($(\bm{s}, \bm{o})$) with the corresponding predicates (\ie, $\mathcal{F}(\bm{s}, \bm{o}) \approx \bm{p}$) in the common embedding space for relation recognition.
To achieve that, we propose a Prototype-guided Learning (PL), to help PE-Net learn the entity-predicate matching. 
Furthermore, to relieve the ambiguous matching problem caused by the predicate's semantic overlap, Prototype Regularization (PR) is proposed to encourage inter-class distinction for accurate entity-predicate matching.

\subsection{Prototype-based Embedding Network}
\label{subsec:network}
The procedure of Prototype-based Embedding Network (PE-Net) can be divided into two steps: 1) Prototype-based Modeling for producing compact and distinctive entity and predicate representations. 2) Prototype-guided Entity-Predicate Matching for relation recognition.

\begin{figure}[!t]
  \center
  \includegraphics[width=0.77\linewidth]{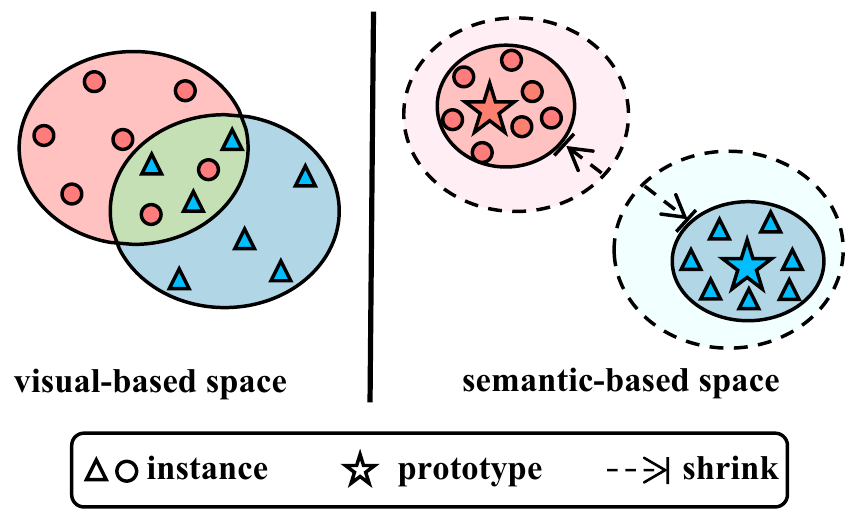}
\caption{The illustration of Prototype-based Modeling \vs Visual-based Modeling. \textbf{Left:} Modeling instances with the appearance in visual-based space suffers from large intra-class variation and severe inter-class overlap.
\textbf{Right:} Our Prototype-based Modeling method gathers instances around their semantic prototypes in the semantic-based space producing compact and distinctive representations. 
}
\label{fig:modeling}
\end{figure}

\noindent \textbf{Prototype-based Modeling.}
The Prototype-based Embedding Network (PE-Net) models entity/predicate instances with prototype-based compact and discriminative representations shown in Fig.~\ref{fig:modeling}.

Concretely, the representations of subject ($\bm s$), object ($\bm o$), and predicate ($\bm p$) are modeled below:
\begin{equation}
\begin{aligned}
    \bm{s} & = \mathbf{W}_s \bm{t_s + v_s}, \\
    \bm{o} & = \mathbf{W}_o \bm{t_o + v_o}, \\
    \bm{p} & = \mathbf{W}_p \bm{t_p + u_p}, 
\label{eq:model}
\end{aligned}
\end{equation}
where $\mathbf{W_s}$, $\mathbf{W_o}$, and $\mathbf{W_p}$ are learnable parameters.
Moreover, $\mathbf{W}_s \bm{t_s}$, $\mathbf{W}_o \bm{t_o}$ and $\mathbf{W}_p \bm{t_p}$ are the class-specific semantic prototypes obtained from their class labels' word embedding (GloVe~\cite{glove}), \ie, $\bm{t_s}$, $\bm{t_o}$ and $\bm{t_p}$. 
Based on the class-specific prototypes, the instance-varied semantic contents $\bm{v_s}$, $\bm{v_o}$ and $\bm{u_p}$ are utilized to model the diversity of each instance from the same subject, object, and predicate class.
Practically,  $\bm{v_s}$ are obtained as:
\begin{equation}
\begin{aligned}
    \bm{g_s} & = \sigma\left(f(\left(\mathbf{W_s}\bm{t_s}) \oplus h(\bm{x_s})\right)\right) \\ 
    \bm{v_s} & = \bm{g_s} \odot h(\bm{x_s}), 
\end{aligned}
\label{eq:visual-content}
\end{equation}
where $f(\cdot)$ is a fully connected layer, $h(\cdot)$ is the visual-to-semantic function used to transform the visual feature into semantic space, and $\oplus$ is the concatenation operation. Moreover, $\sigma(\cdot)$ is the sigmoid activation function, $\odot$ is the element-wise product, and \bm{$x_s$} is the visual features of subject instances from the detector.
Utilizing the gate mechanism in \cref{eq:visual-content}, the class-irrelevant information is eliminated from the original visual feature \bm{$x_s$} producing consistent representations within class. In addition, we derive $\bm{v_o}$ in the same way following \cref{eq:visual-content}. 

Similarly, predicate's instance-varied semantic content $\bm{u_p}$ is defined as:
\begin{equation}
\label{p}
    \begin{aligned}
    \bm{g_p} & = \sigma \left( f\left( \mathcal{F}(\bm{s,o}) \oplus h(\bm{x_u})  \right)  \right) \\ 
    \bm{u_p} & = \bm{g_p} \odot h(\bm{x_u}), 
\end{aligned}
\end{equation}
where $\bm{x_u}$ is the union feature of subject and object, and $\mathcal{F}(\cdot, \cdot)$ denotes the feature fusion function. 



\noindent \textbf{Prototype-guided Entity-Predicate Matching.}
Then, for relation recognition, we match subject instance ($\bm s$) and object instance ($\bm o$) with the corresponding predicate instance ($\bm p$) in the common semantic space.
Practically, the entity-predicate matching is shown below:
\begin{equation}
\label{eq:sor1}
    \begin{aligned}
   \mathcal{F}(\bm{s}, \bm{o}) \approx  \bm{p},
   \end{aligned}
\end{equation}
where $\mathcal{F}(\bm{s}, \bm{o})$ is defined as: $\text{ReLU}(\bm{s+o}) - (\bm{s-o})^2$.

However, the predicate representation varies with the subject-object pair, which prevents PE-Net from efficiently learning the matching.
Therefore, we perform an equivalent transformation on \cref{eq:sor1}, deriving a deterministic matching objective as follows:
\begin{equation}
\label{eq:sor2}
    \mathcal{F}(\bm{s}, \bm{o}) - \bm{u_p} \approx \mathbf{W}_p \bm{t_p},
\end{equation}
where $\mathcal{F}(\bm{s}, \bm{o}) - \bm{u_p}$ is defined as relation representation $\bm{r}$, which should be matched to its corresponding predicate prototype $\mathbf{W}_p \bm{t_p}$ (represented as \bm{$c$} in the following sections).

\subsection{Prototype-guided Learning}
\label{subsec:constraint}
To help PE-Net efficiently match relation representations with corresponding predicates in \cref{eq:sor2}, we devise a learning strategy, \ie, Prototype-guided Learning (PL), which makes relation representations close to their corresponding prototypes. In practice, PL consists of two constraints: cosine similarity and Euclidean distance.

Firstly, we have to increase the cosine similarity between the relation representation \bm{$r$} and its corresponding prototype \bm{$c_t$}, which is implemented as the following loss function:
\begin{equation}
\label{eq:loss_cos}
  \mathcal{L}_{e\_sim} =  -   \log{\frac{\exp(\langle\bm{ \overline{r}, \overline{c_t}} \rangle / \tau)}{\sum_{j=0}^{N} \exp(\langle\bm{ \overline{r}, \overline{c_j}} \rangle / \tau)}},
\end{equation}
where \bm{$\overline{\cdot}$} denotes the unitary operation, 
$\tau$ is a learnable temperature hyper-parameter, $t$ is subscript for the ground truth class, and $N$ is the number of predicate categories. 

Since the cosine similarity only considers angle-based relative distance, it may fail to make relation representations and corresponding prototypes close to each other in the Euclidean space.
To the end, we further impose the Euclidean distance constraint.
It encourages the relation representation \bm{$r$} close to its corresponding prototypes \bm{$c_t$} while keeping the distances with others in Euclidean space.
Practically, we first calculate the distances between the relation representation \bm{$r$} and each class prototype \bm{$c_i$} obtaining the distances set $\bm{G} = \{ g_i\}_{i=0}^N$, where $g_i$ is computed as:
\begin{equation}
    g_i = {\|\bm{r} - \bm{c_i}\|}_2^2.
\end{equation}
Then, we sort the distance set $\bm{B} = \bm{G} \setminus \{g_t\}$ (excluding $g_t$) in increasing order and obtain the sorted distance set as $\bm{B'} = {\{ {b'_i}\} _{i = 0}^{N-1}}$.
Furthermore, the top $k_1$ smallest distances of $\bm{B'}$ are averaged as the distance $g^-$ to negative prototypes:
\begin{equation}
    g^- = \frac{1}{k_1} \sum_{i=0}^{k_1-1} b'_i.
\end{equation}
Together with the distance to the positive prototype $g^+ = g_t$, we further construct the triplet loss:
\begin{equation}
    \mathcal{L}_{e\_euc} = \max{(0, g^+ - g^- + \gamma_1)},
\end{equation}
where $\gamma_1$ is a hyper-parameter to adjust the distance margins between relation representations and the negative prototypes.

\subsection{Prototype Regularization}
To alleviate the ambiguous matching caused by the semantic overlap between predicates, we propose a Prototype Regularization (PR) to encourage inter-class separation by enlarging the distinction between prototypes for precise entity-predicate matching.
Correspondingly, according to the constraints imposed in \cref{subsec:constraint}, we first calculate the cosine similarity between predicate prototypes obtaining the similarity matrix as follows:
\begin{equation}
    \bm{S} = \bm{\overline{C}} \cdot \bm{\overline{C}^{\top}} = (s_{ij}) \in \mathbb{R}^{(N+1) \times (N+1)},
\end{equation}
where $\bm{C}=[\bm{c_0}; \bm{c_1};...; \bm{c_N}]$ is the predicate prototype matrix, and
\bm{$\overline{C}$} is obtained by normalizing the vectors in it. Moreover, $s_{ij}$ represents the cosine similarity between prototype $\bm{c_i}$ and $\bm{c_j}$.
Then, we should reduce each pair of prototypes' cosine similarity to make them distinctive in the semantic space.
Therefore, we introduce the $l_{2,1}$-norm of \bm{$S$} and minimize it:
\begin{equation}
    \mathcal{L}_{r\_sim} = {\|\bm{S}\|}_{2,1}= \sum_{i=0}^{N} \sqrt{\sum_{j=0}^{N} s_{ij}^2}.
\end{equation}

However, only regularized by the cosine similarity, some predicates are still not distinctive enough against others. Thus, we enlarge their distances in the Euclidean space for further distinction. 
To achieve that, we calculate the Euclidean distance between two prototypes obtaining the distance matrix 
$\bm{D} = (d_{ij}) \in \mathbb{R}^{(N+1) \times (N+1)}$ with $d_{ij}$ computed as:
\begin{equation}
    d_{ij} = {\|\bm{c_i} - \bm{c_j} \|}_2^2, 
\end{equation}
where $ d_{ij}$ indicates the Euclidean distance between prototype $\bm{c_i}$ and $\bm{c_j}$.
For each prototype, we should distance it from others. 
Therefore, we sort the elements in each row of matrix \bm{$D$} in increasing order, obtaining $\bm{D'} =  (d'_{ij}) \in \mathbb{R}^{(N+1) \times (N+1)}$, and select the top $k_2$ smallest values of each row to widen them:
\begin{equation}
\begin{aligned}
    d^- & = \frac{1}{(N+1)k_2} \sum_{i=0}^{N} \sum_{j=1}^{k_2} d'_{ij}, \\
    \mathcal{L}_{r\_euc} & = \max (0, - d^- + \gamma_2), 
\end{aligned}
\end{equation}
where $\gamma_2$ is another hyper-parameter used to adjust the distance margins.

\subsection{Scene Graph Prediction}
\label{subces:inference}
During the training stage, the final loss function $\mathcal{L}$ for our PE-Net is expressed as:
\begin{equation}
\label{loss}
    \mathcal{L} = \mathcal{L}_{r\_sim} + \mathcal{L}_{e\_sim} + \mathcal{L}_{r\_euc} + \mathcal{L}_{e\_euc} .  
\end{equation}

During the testing stage, with the relation representation \bm{$r$}, we choose the class of prototypes with the highest cosine similarity as the prediction result: 
\begin{equation}
    res_{r} = \underset{i}{ \operatorname {arg\,max} } (\{q_i | q_i = \langle \bm{\overline{r}}, \bm{\overline{c_i}} \rangle / \tau \}),
\end{equation}
where $q_i$ indicates the similarity between relation representation \bm{$r$} and prototype \bm{$c_i$}.

\section{Experiments}
\label{sec:experiments}

\begin{table*}[htp]
\centering 
\resizebox{\textwidth}{!}{
\begin{tabular}{c|ccc|ccc|ccc}
\hline
\multirow{2}{*}{Model} & \multicolumn{3}{c|}{PredCls} & \multicolumn{3}{c|}{SGCls} & \multicolumn{3}{c}{SGDet} \\
\cline{2-10}
& R@50/100 & mR@50/100 & M@50/100 & R@50/100 & mR@50/100 & M@50/100 & R@50/100 & mR@50/100 & M@50/100\\
\hline
SGTR~\cite{sgg:sgtr} & - & - & - & - & - & - & 24.6 / 28.4 & 12.0 / 15.2 & 18.3 / 21.8 \\ 
SS R-CNN~\cite{sgg:ssrcnnn} & - & - & - & - & - & - & 33.5 / 38.4 & 8.6 / 10.3 & 21.1 / 24.4 \\ 
\hline
Motifs$^{\diamond}$~\cite{sgg:motifs,sgg:sggbenchmark} & 65.3 / 67.2 & 14.9 / 16.3 & 40.1 / 41.8 & 38.9 / 39.8 & 8.3 / 8.8 & 23.6 / 24.3 & 32.1 / 36.8 & 6.6 / 7.9 & 19.4 / 22.4 \\
VCTree$^{\diamond}$~\cite{sgg:vctree,sgg:sggbenchmark} & 65.5 / 67.4 & 16.7 / 17.9 & 41.1 / 42.7 & 40.3 / 41.6 & 7.9 / 8.3 & 24.1 / 25.0 & 31.9 / 36.0 & 6.4 / 7.3 & 19.2 / 21.7 \\
G R-CNN$^{\star}$~\cite{sgg:graphrcnn, sgg:bgnn} & 65.4 / 67.2 & 16.4 / 17.2 & 40.9 / 42.2 & 37.0 / 38.5 & 9.0 / 9.5 & 23.0 / 24.0 & 29.7 / 32.8 & 5.8 / 6.6 & 17.8 / 19.7 \\
KERN$^{\star}$~\cite{sgg:kern,sgg:bgnn} & 65.8 / 67.6 & 17.7 / 19.2 & 41.8 / 43.4 & 36.7 / 37.4 & 9.4 / 10.0 & 23.1 / 23.7 & 27.1 / 29.8 & 6.4 / 7.3 &  16.8 / 18.6\\
VTransE$^{\diamond}$~\cite{sgg:vtranse,sgg:sggbenchmark} & 65.7 / 67.6 & 14.7 / 15.8 & 40.2 / 41.7 & 38.6 / 39.4 & 8.2 / 8.7 & 23.4 / 24.1 & 29.7 / 34.3 & 5.0 / 6.1 & 17.4 / 20.2 \\
R-CAGCN~\cite{sgg:pum} & 66.6 / 68.3 & 18.3 / 19.9 & 42.5 / 44.1 & 38.3 / 39.0 & 10.2 / 11.1 & 24.3 / 25.1 & 28.1 / 31.3 & 7.9 / 8.8 & 18.0 / 20.1 \\ 
GPS-Net$^{\star}$~\cite{sgg:gps,sgg:bgnn} & 65.2 / 67.1 & 15.2 / 16.6 & 40.2 / 41.9 & 37.8 / 39.2 & 8.5 / 9.1 & 23.2 / 24.2 & 31.3 / 35.9 & 6.7 / 8.6 & 19.0 / 22.3 \\ 
RU-Net~\cite{sgg:runet} & \underline{67.7} / \underline{69.6} & - / 24.2 & - / 46.9 & \textbf{42.4} / \textbf{43.3} & - / 14.6 & - / \underline{29.0} & \textbf{32.9} / \textbf{37.5} & - / 10.8 & - / \underline{24.2} \\ 
BGNN$^{\star}$~\cite{sgg:bgnn} & 59.2 / 61.3 & \underline{30.4} / \underline{32.9} & 44.8 / 47.1 & 37.4 / 38.5 & \underline{14.3} / \underline{16.5} & 25.9 / 27.5 & 31.0 / 35.8 & \underline{10.7} / \underline{12.6} & \underline{20.9} / \underline{24.2} \\
\hline
\textbf{PE-Net(P)} & \textbf{68.2} / \textbf{70.1} & 23.1 / 25.4 & \underline{45.7} / \underline{47.8} & \underline{41.3} / \underline{42.3} & 13.1 / 14.8 & \underline{27.2} / 28.6 & \underline{32.4} / \underline{36.9} & 8.9 / 11.0 & 20.7 / 24.0 \\ 
\textbf{PE-Net} & 64.9 / 67.2 & \textbf{31.5} / \textbf{33.8} & \textbf{48.2} / \textbf{50.5} & 39.4 / 40.7 & \textbf{17.8} / \textbf{18.9} & 
\textbf{28.6} / \textbf{29.8} & 30.7 / 35.2 & \textbf{12.4} / \textbf{14.5} & \textbf{21.6} / \textbf{24.9} \\ 
\hline
\hline

Motifs-TDE~\cite{sgg:tde} & 46.2 / 51.4 & 25.5 / 29.1 & 35.9 / 40.3 & 27.7 / 29.9 & 13.1 / 14.9 & 20.4 / 22.4 & 16.9 / 20.3 & 8.2 / 9.8 & 12.6 / 15.1 \\ 
Motifs-CogTree~\cite{sgg:cogtree} & 35.6 / 36.8 & 26.4 / 29.0 & 31.0 / 32.9 & 21.6 / 22.2 & 14.9 / 16.1 & 18.3 / 19.2 & 20.0 / 22.1 & 10.4 / 11.8 & 15.2 / 17.0 \\ 
Motifs-BPL-SA~\cite{sgg:ba-sgg} & 50.7 / 52.5 & 29.7 / 31.7 & 40.2 / 42.1 & 30.1 / 31.0 & 16.5 / 17.5 & 23.3 / 24.3 & 23.0 / 26.9 & 13.5 / 15.6 & 18.3 / 21.3 \\
Motifs-NICE~\cite{sgg:nice} & \underline{55.1} / \underline{57.2} & 29.9 / 32.3 & 42.5 / 44.8 & \underline{33.1} / \underline{34.0} & 16.6 / 17.9 & \underline{24.9} / 26.0 & \textbf{27.8} / \textbf{31.8} & 12.2 / 14.4 & \underline{20.0} / \underline{23.1} \\ 
Motifs-PPDL~\cite{sgg:ppdl} & 47.2 / 47.6 & 32.2 / 33.3 & 39.7 / 40.5 & 28.4 / 29.3 & 17.5 / 18.2 & 23.0 / 23.8 & 21.2 / 23.9 & 11.4 / 13.5 & 16.3 / 18.7 \\
Motifs-GCL~\cite{sgg:gcl} & 42.7 / 44.4 & \underline{36.1} / \underline{38.2} & 39.4 / 41.3 & 26.1 / 27.1 & \underline{20.8} / \underline{21.8} & 23.5 / 24.5 & 18.4 / 22.0 & \textbf{16.8} / \textbf{19.3} & 17.6 / 20.7 \\ 
Motifs-Reweight~\cite{loss:cbloss} & 53.2 / 55.5 & 33.7 / 36.1 & \underline{43.5} / \underline{45.8}  & 32.1 / 33.4 & 17.7 / 19.1 & \underline{24.9} / \underline{26.3} & 25.1 / 28.2 & 13.3 / 15.4 & 19.2 / 21.8 \\ 
\hline
\textbf{PE-Net-Reweight} & \textbf{59.0} / \textbf{61.4} & \textbf{38.8} / \textbf{40.7} & \textbf{48.9} / \textbf{51.1} & \textbf{36.1} / \textbf{37.3} & \textbf{22.2} / \textbf{23.5} & \textbf{29.2} / \textbf{30.4} & \underline{26.5} / \underline{30.9} & \underline{16.7} / \underline{18.8} & \textbf{21.6} / \textbf{24.9} \\ 
\hline
\end{tabular}
}
\caption{Performance comparison with the state-of-the-art SGG methods on VG dataset. 
$^{\star}$ and $^{\diamond}$ denotes the results reproduced with the codebase provided by~\cite{sgg:bgnn} and ~\cite{sgg:sggbenchmark}. 
\textbf{PE-Net(P)} refers to the PE-Net only trained with PL. \textbf{PE-Net} indicates PE-Net trained with both PL and PR.
The \textbf{best} and \underline{second best} methods under each setting are marked according to formats. }
\label{tab:res_vanilla}
\end{table*}

\subsection{Datasets}
\label{subsec:dataset-setting}
\noindent \textbf{Visual Genome (VG). } The Visual Genome (VG) dataset consists of 108,077 images with average annotations of 38 objects and 22 relationships per image. In this paper, we adopt the most widely used split~\cite{sgg:imp}, which contains the most frequent 150 object categories and 50 predicate categories.
Specifically, the dataset is divided into a training set with 70\% of the images, a testing set with the remaining 30\%, and 5$k$ images from the training set for validation. 

\noindent \textbf{Open Images (OI).}
We conduct experiments on Open Image V6 dataset, which has 126,368 images for training, and 1813 and 5322 images for validation and testing. It contains 301 object categories and 31 predicate categories.

\subsection{Evaluation Protocol}
\noindent \textbf{Visual Genome (VG).}
We evaluate our method on three sub-tasks, including Predicate Classification (\textbf{PredCls}), Scene Graph Classification (\textbf{SGCls}), and Scene Graph Detection (\textbf{SGDet}).
Following the recent works~\cite{sgg:tde,sgg:sggbenchmark,sgg:vctree,sgg:fgpl}, we take Recall@K (\textbf{R@K}) and mean Recall@K (\textbf{mR@K}) as the primary evaluation metrics.
Moreover, we also report the zero-shot Recall@K (\textbf{zs-R@K}) that measures the model's generalization in dealing with the unseen relation triplets during training. 
Due to the imbalanced data distribution of VG dataset, R@K focuses on the common predicates with abundant samples, and mR@K prefers the tail predicates.
Therefore, we introduce the Mean@K (\textbf{M@K}), which averages the R@K and mR@K for evaluating the model's overall performance on SGG.
In addition, the Intra-class Variance (\textbf{IV}) and Intra-class to Inter-class Variance Ratio (\textbf{IIVR}) are introduced to measure the compactness and distinctiveness of entity/predicate representations. Intuitively, lower values of \textbf{IV} and \textbf{IIVR} indicate higher quality for representations. 


\noindent \textbf{Open Images (OI).}
Following the previous works~\cite{sgg:bgnn,sgg:runet,sgg:reldn}, we utilize the Recall@50 (\textbf{R@50}), weighted mean AP of relations (\textbf{wmAP$_{rel}$}), weighted mean AP of phrase (\textbf{wmAP$_{phr}$}) as the evaluation metrics. The \textbf{score$_{wtd}$} is calculated as:
$\text{score}_{wtd} = 0.2 \times \text{R@50} + 0.4 \times \text{wmAP}_{rel} + 0.4 \times \text{wmAP}_{phr}$. 


\subsection{Implementation Details}
\label{subsec:implement-detail}
Following the previous works~\cite{sgg:sggbenchmark,sgg:tde,sgg:hlnet,sgg:pum}, we adopt the Faster R-CNN~\cite{fasterrcnn} with ResNeXt-101-FPN~\cite{fpn,resxnet,resnet} provided by~\cite{sgg:sggbenchmark} to detect objects in the image. The parameters of the detector are frozen during training. 
In particular, we set $k_1$, $k_2$, $\gamma_1$, and $\gamma_2$ as 10, 1, 1, and 7.
Additionally, the PE-Net is trained by an SGD optimizer with 60$k$ iterations. The initial learning rate and the batch size are set to $10^{-3}$ and 8. All experiments are implemented with PyTorch and trained with an NVIDIA GeForce RTX 3090 GPU. 

\subsection{Comparisons with State-of-the-art Methods}
\label{subsec:compare_sota}

\noindent \textbf{Visual Genome.}
To evaluate PE-Net's capability on scene graph generation, we compare it with several state-of-the-art SGG methods on Visual Genome dataset under all three sub-tasks. The results are shown in \cref{tab:res_vanilla}. 
%
%
Generally, our method achieves superior performance compared to other SGG models.
Concretely, PE-Net(P) outperforms the VCTree by 7.5\%, 6.5\%, and 3.7\% at mR@100 and by 2.7\%, 0.7\%, and 0.9\% at R@100 on PredCls, SGCls, and SGDet.
It also outperforms the recent SGG model, RU-Net, by 0.5\% and 1.2\% at R@100 and mR@100 on PredCls.
In addition, the full PE-Net outperforms 
VCTree by 15.9\%, 10.6\%, 7.2\%,  
and RU-Net by 9.6\%, 4.3\%, 3.7\%,
at mR@100 on three subtasks, respectively. 
The results demonstrate the effectiveness of our model.


Moreover, to explore PE-Net's potential capability of solving the long-tail problem for SGG, we equip it with the advanced re-weighting method~\cite{loss:cbloss}.
Then, we compare PE-Net-Reweight with Motifs~\cite{sgg:motifs} de-biased by several existing state-of-the-art de-biasing methods.
The results are summarized in \cref{tab:res_vanilla}.
We find that our PE-Net-Reweight pushes the performance on unbiased SGG to a new level.
For instance, compared with Motifs-GCL, we achieve an absolute performance advantage, outperforming it by 17.0\%, 10.2\%, and 8.9\% at R@100 on PredCls, SGCls, and SGDet tasks, and by 2.5\%, 1.7\% at mR@100 on PredCls, SGCls tasks, respectively.
Benefiting from the prototype-aligned distinctive representations, the PE-Net has the potential to tackle the biased problem in SGG.

In addition, we report the zero-shot recall results to verify the generalization of our method in handling the unseen relation triplets in the training set. 
As shown in \cref{tab:zs-r}, our model outperforms the vanilla Motifs and VCTree by 15.53\%, 5.40\%, 3.49\%, and 15.38\%, 4.45\%, 2.91\% at zs-R@100 on PredCls, SGCls, and SGDet. 
Although TDE significantly improves the zero-shot performance by removing the effect of context bias, Motifs-TDE and VCTree-TDE are still surpassed by our PE-Net with 2.69\%, 2.03\%, 0.7\%, and 3.29\%, 2.53\%, 0.4\% on three tasks, respectively. 
We owe the strength to the Prototype-based Modeling of our PE-Net, which models entity and predicate in the semantic space, significantly improving the model's analogical reasoning capability on unseen relation triplets.

\noindent \textbf{Open Images.}
To verify the generality of our method on different datasets, we conduct experiments on Open Images and present the results in \cref{tab:oi}. 
Consistent with the performance on VG, PE-Net also achieves competitive results on Open Images dataset.
Specifically, our method exceeds the BGNN with a large margin of 2.7\% on average at four metrics, and outperforms RU-Net by 1.2\%, 2.5\%, and 1.4\% at wmAP$_{rel}$, wmAP$_{phr}$, and score$_{wtd}$, respectively. 
It powerfully confirms PE-Net's generalization on handling relation recognition under different data distributions.

\begin{table}[!t] \centering 
\resizebox{\linewidth}{!}{
    \begin{tabular}{c|ccc}
    \hline
    \multirow{2}{*}{Models} & PredCls & SGCls & SGDet   \\
    \cline{2-4}
    & zs-R@50 / 100 & zs-R@50 / 100 & zs-R@50 / 100  \\
    \hline
    Motifs~\cite{sgg:motifs}& 3.24 / 5.36 & 0.68 / 1.13 & 0.05 / 0.11 \\  
    VCTree~\cite{sgg:vctree}& 3.27 / 5.51 & 1.17 / 2.08 & 0.31 / 0.69 \\ 
    Motifs-TDE~\cite{sgg:tde} & 14.4 / 18.2 & 3.4 / 4.5 & 2.3 / 2.9 \\  
    VCTree-TDE~\cite{sgg:tde} & 14.3 / 17.6 & 3.2 / 4.0 & \textbf{2.6} / 3.2 \\
    Motifs-EBM~\cite{sgg:ebm} & 4.87 / - & 1.25 / - & 0.23 / - \\ 
    VCTree-EBM~\cite{sgg:ebm} & 5.36 / - & 1.87 / - & 0.54 / - \\ 
    \hline
    \textbf{PE-Net} & \textbf{17.16} / \textbf{20.89} & \textbf{5.37} / \textbf{6.53} & 2.31 / \textbf{3.60} \\  
    \hline
    \end{tabular}
}
\caption{Comparison with different methods on Zero-shot Recall (zs-R@50/100) under all three sub-tasks on the VG dataset.}
\label{tab:zs-r} 
\end{table}

\begin{table}[!t] \centering 
\resizebox{0.9\linewidth}{!}{
    \begin{tabular}{c|cccc}
    \hline
    Model & R@50 & wmAP$_{rel}$ & wmAP$_{phr}$ & score$_{wtd}$ \\
    \hline
    Motifs~\cite{sgg:motifs} & 71.6 & 29.9 & 31.6 & 38.9 \\
    G R-CNN~\cite{sgg:graphrcnn} & 74.5 & 33.2 & 34.2 & 41.8 \\ 
    GPS-Net~\cite{sgg:gps} & 74.8 & 32.9 & 34.0 & 41.7 \\
    VCTree~\cite{sgg:vctree} & 74.1 & 34.2 & 33.1 & 40.2 \\ 
    BGNN~\cite{sgg:bgnn} & 75.0 & 33.5 & 34.2 & 42.1 \\ 
    RU-Net~\cite{sgg:runet} & \textbf{76.9} & \underline{35.4} & \underline{34.9} & \underline{43.5} \\ 
    \hline
    \textbf{PE-Net} & \underline{76.5} & \textbf{36.6} & \textbf{37.4} & \textbf{44.9} \\
    \hline
    \end{tabular}
}
\caption{Comparison with the state-of-the-art methods on Open-Images V6. We adopt the same evaluation metric as in \cite{sgg:bgnn}. The \textbf{best} and \underline{second best} methods under each setting are marked according to formats. }
\label{tab:oi} 
\end{table}

\begin{table*}[htp] \centering 
\resizebox{\textwidth}{!}{
\begin{tabular}{c|cc|ccc|ccc|ccc}
\hline

\multirow{2}{*}{Exp} & \multicolumn{2}{c|}{Component} & \multicolumn{3}{c|}{PredCls} & \multicolumn{3}{c|}{SGCls} & \multicolumn{3}{c}{SGDet} \\
\cline{2-12}
& PL & PR & R@50/100 & mR@50/100 & M@50/100 & R@50/100 & mR@50/100 & M@50/100 & R@50/100 & mR@50/100 & M@50/100 \\
\hline
1 & \ding{55} & \ding{55} &
66.5 / 68.2 & 18.5 / 20.0 & 42.5 / 44.1 & 39.5 / 40.4 & 9.9 / 10.5 & 24.7 / 25.5  & 32.3 / 36.8 & 7.8 / 9.3 & 20.1 / 23.1 \\ 
2 & \ding{51} & \ding{55} & \textbf{68.2} / \textbf{70.1} & 23.1 / 25.4 & 45.7 / 47.8 & \textbf{41.3} / \textbf{42.3} & 13.1 / 14.8 & 27.2/ 28.6 & \textbf{32.4} / \textbf{36.9} & 8.9 / 11.0 & 20.7 / 24.0 \\
3  & \ding{51} & \ding{51} &
64.9 / 67.2 & \textbf{31.5} / \textbf{33.8} & \textbf{48.2} / \textbf{50.5} & 39.4 / 40.7 & \textbf{17.8} / \textbf{18.9} & \textbf{28.6} / \textbf{29.8} & 30.7 / 35.2 & \textbf{12.4} / \textbf{14.5} & \textbf{21.6} / \textbf{24.9} \\
\hline
\end{tabular}
}
\caption{Ablation study on each component of PE-Net. PL and PR denote the Prototype-guided Learning and Prototype Regularization. }

\label{tab:ablation_study}
\end{table*}

\begin{figure}[!t]
\centering
\subfloat[\small{Entities (Motifs)}]{\includegraphics[width=1.5in]{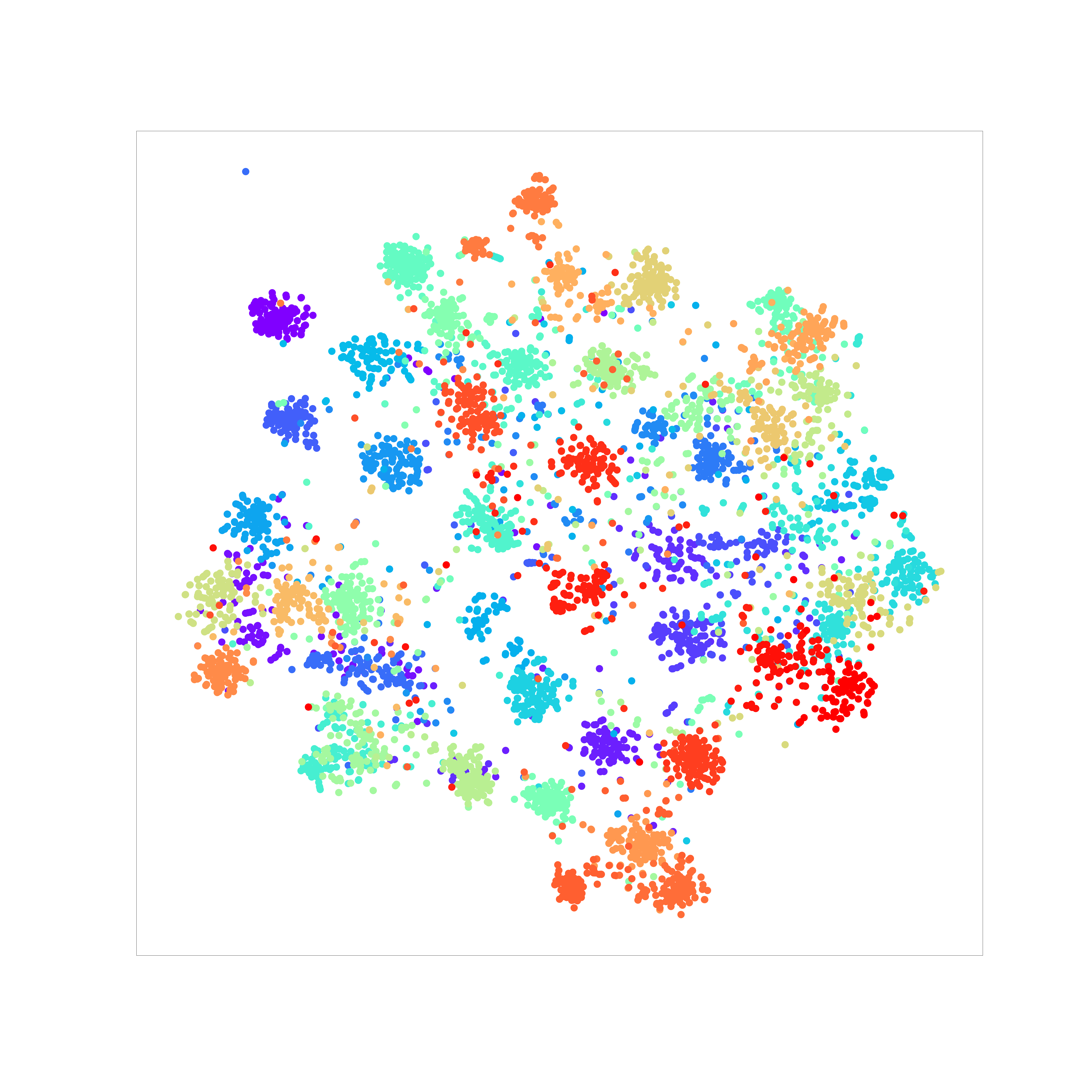}%
\label{fig:feature_distaibution_a}}
\hfil
\subfloat[\small{Entities (PE-Net)}]{\includegraphics[width=1.5in]{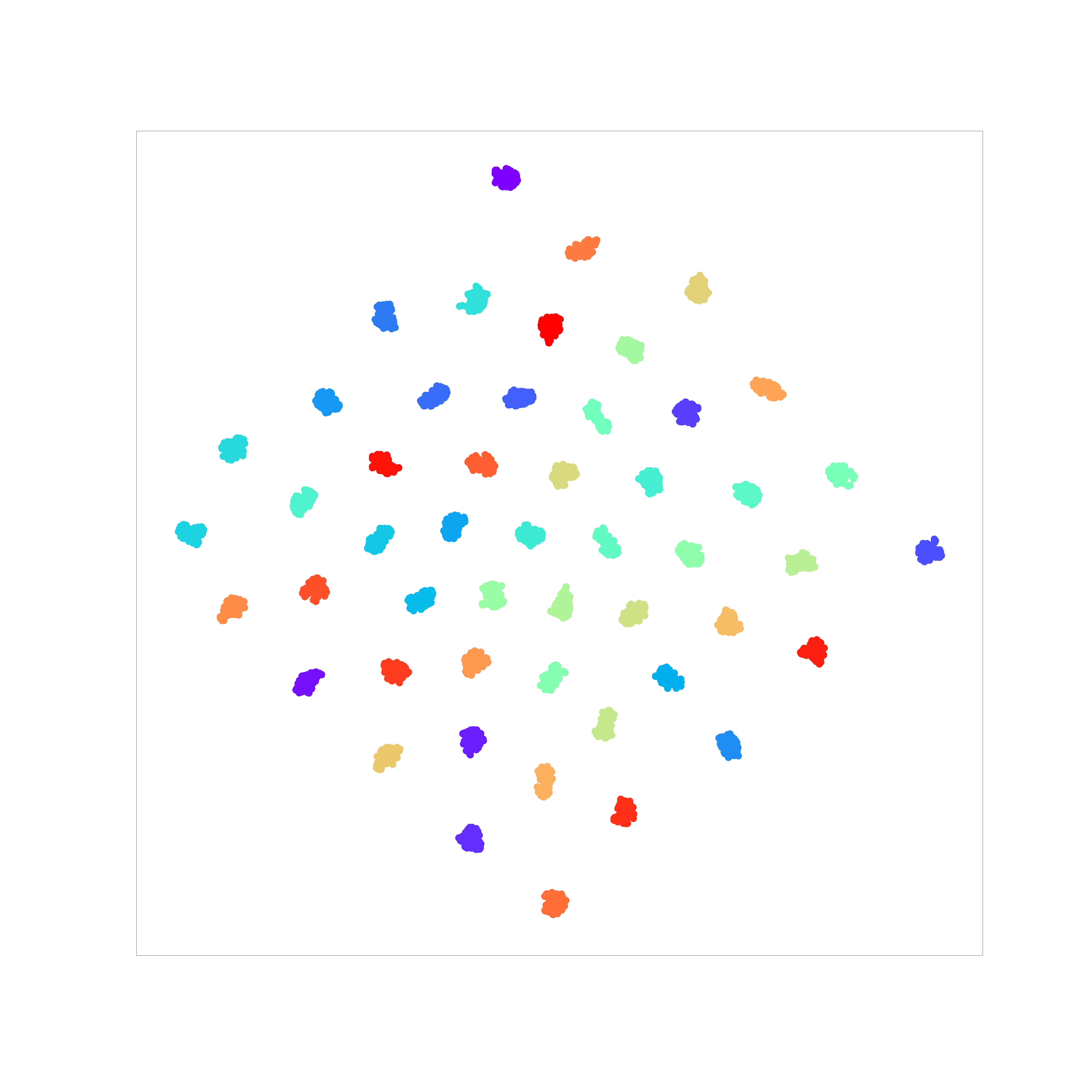}%
\label{fig:feature_distaibution_b}}

\subfloat[\small{Relations (Motifs)}]{\includegraphics[width=1.5in]{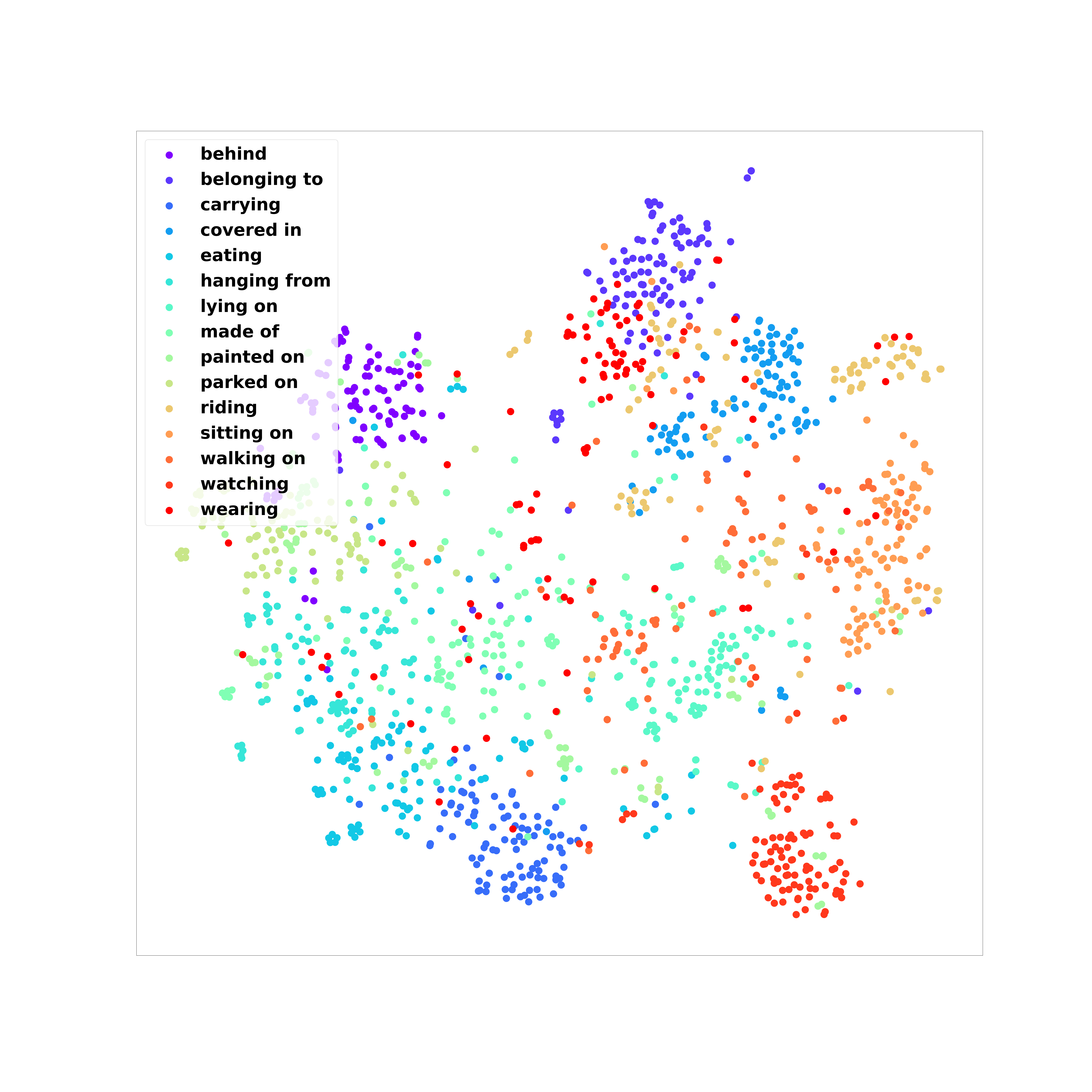}%
\label{fig:feature_distaibution_c}}
\hfil
\subfloat[\small{Relations (PE-Net)}]{\includegraphics[width=1.5in]{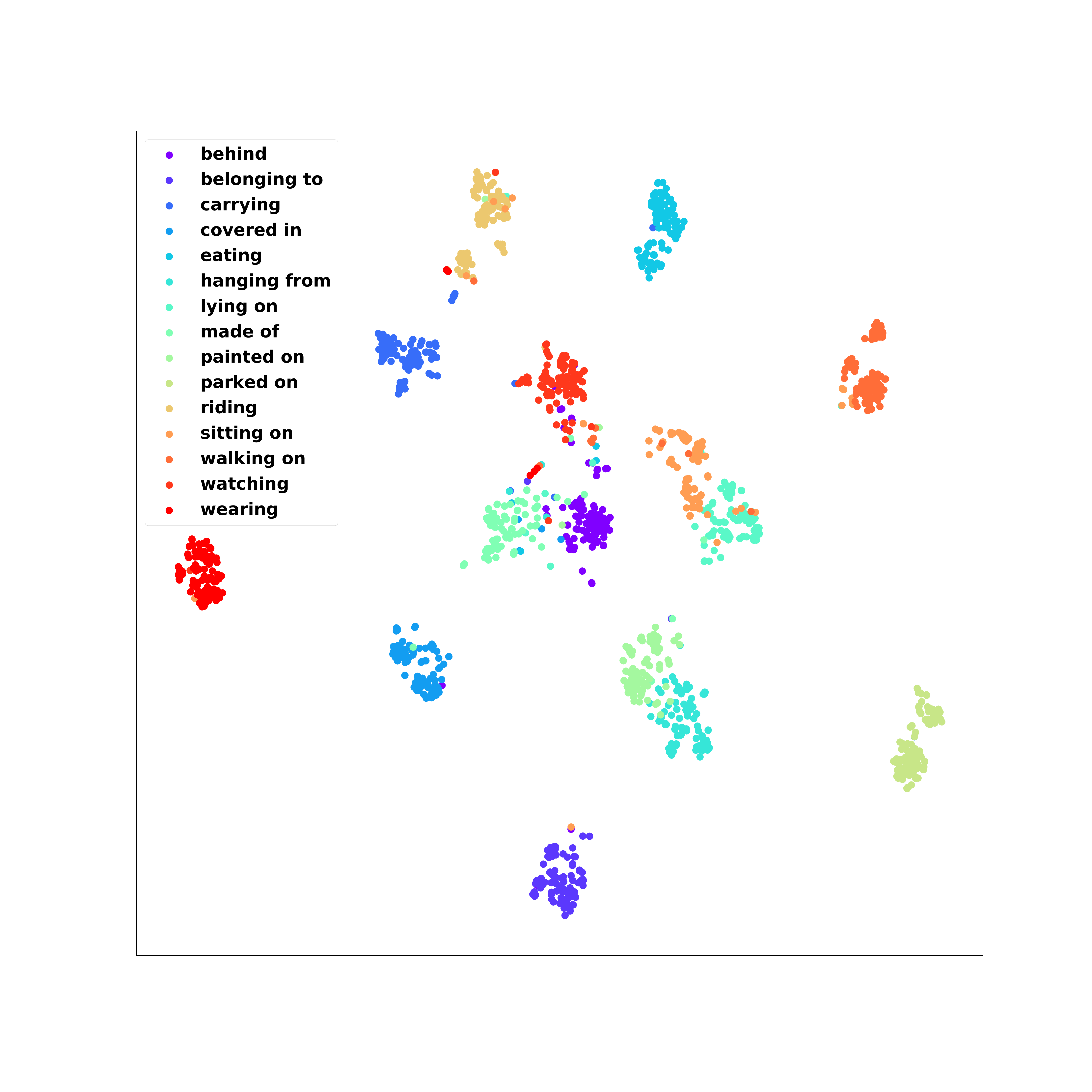}%
\label{fig:feature_distaibution_d}}
\caption{
The comparison of t-SNE visualization results on entity and predicate feature distributions between PE-Net and Motifs under the VG dataset. 
}
\label{fig:feature_distribution}  
\end{figure}

\subsection{Measuring Representation Modeling of PE-Net}
\label{subsec:analysis}
To certify the assumption that our PE-Net is capable of producing compact and distinctive representations for entity and predicate, we conduct both quantitative and qualitative studies in \cref{tab:iv} and \cref{fig:feature_distribution}, respectively.
Notably, we only conduct experiments on the PredCls task, which eliminates the impact of entities' mis-classification made by detectors. 

\noindent \textbf{Quantitative Analysis}.
To quantitatively evaluate the quality of the entity's and predicate's representations (\ie, degree of intra-class compactness and the inter-class distinctiveness), we evaluate and make comparisons between PE-Net and previous methods~\cite{sgg:motifs,sgg:vctree,sgg:graphrcnn,sgg:gps,sgg:sggbenchmark} with IV (Intra-class Variance) and IIVR (Intra-class to Inter-class Variance Ratio). Moreover, the experimental results are shown in \cref{tab:iv}.
%
Firstly, our PE-Net yields more compact entity and predicate representations than previous methods, \eg, 0.74 \vs 9.73 on IV-O and 1.06 \vs 1.41 on IV-R compared with Motifs. 
That illustrates the effectiveness of our Prototype-based Modeling in PE-Net. 
Also, the representations learned by our model are more distinguishable, \eg, 0.24 \vs 1.93 on IIVR-O and 1.67 \vs 2.72 on IIVR-R compared with Motifs. 
We owe it to the effectiveness of our PR, which significantly alleviates the ambiguous entity-predicate matching by encouraging predicate prototypes away from each other.

\noindent \textbf{Qualitative Analysis}.
For an intuitive illustration of PE-Net's capability of yielding compact and distinguishable representations, we visualize the feature distribution of entities and predicates taking advantage of the t-SNE technique, shown in \cref{fig:feature_distribution}. 
Comparing \cref{fig:feature_distribution}(a) with \cref{fig:feature_distribution}(b), we observe that PE-Net produces more compact and distinctive entity representations than Motifs, intuitively illustrating the advantages of modeling instances in the semantic space than from visual appearances. 
In addition, the relation feature distribution of Motifs shown in \cref{fig:feature_distribution}(c) is of large intra-class variance and severe inter-class overlap. In this case, it is hard for SGG models to learn a perfect decision boundary for accurate relation recognition. 
On the contrary, the relation representations learned by our PE-Net are of high-level inter-class distinctiveness and intra-class compactness, which intuitively demonstrates our method's superiority and explains why our method achieves excellent relation prediction performance.


\begin{table}[!t] \centering 
\resizebox{0.9\linewidth}{!}{
    \begin{tabular}{c|cc|cc}
    \hline
    Models & IV-O \bm{$\downarrow$} & IIVR-O \bm{$\downarrow$} & IV-R \bm{$\downarrow$} & IIVR-R \bm{$\downarrow$} \\ 
    \hline
    Motifs~\cite{sgg:motifs,sgg:sggbenchmark}  & 9.73 & 1.93 & 1.41 & 2.72 \\ 
    VCTree~\cite{sgg:vctree,sgg:sggbenchmark} & 8.31 & 2.11 & 1.50 & 2.78 \\
    Transformer~\cite{transformer,sgg:sggbenchmark} & 9.08 & 2.05 & 1.44 & 2.76 \\
    G-RCNN~\cite{sgg:graphrcnn,sgg:bgnn} & 8.76 & 1.99 & 1.46 & 2.81 \\ 
    GPS-Net~\cite{sgg:gps,sgg:bgnn} & 9.36 & 2.07 & 1.53 & 2.69 \\ 
    \hline
    \textbf{PE-Net}  & \textbf{0.74} & \textbf{0.24} & \textbf{1.06} & \textbf{1.67} \\ 
    \hline
    \end{tabular}
}
\caption{
Quantitative results on representation quality compared with classical SGG models under the PredCls task on VG dataset. The lower values indicate representations with higher quality. 
}
\vspace{-1.2em}
\label{tab:iv} 
\end{table}

\subsection{Ablation Studies}
\label{subsec:ablation_studies}

To verify the effectiveness of each component of the proposed PE-Net, we conduct ablation studies on PL and PR under the VG dataset, and the results are summarized in \cref{tab:ablation_study}. 
Exp 1, PE-Net is trained without PL and PR, which directly uses a linear classifier to classify the relation representation defined in \cref{eq:sor2}. 
Exp 2, PE-Net is trained with PL, which discards the $\mathcal{L}_{r\_sim}$ and $ \mathcal{L}_{r\_euc}$ in \cref{loss}. 
Exp 3, PE-Net is trained with both PL and PR, \ie, \cref{loss}.
When constrained by PL in Exp 2, the model outperforms the baseline (\ie, Exp 1) on all metrics under three sub-tasks (\eg, 25.4\% \vs 20.0\% at mR@100, and 70.1\% \vs 68.2\% at R@100 on PredCls).
This verifies that PL effectively helps PE-Net to establish matching between entities and predicates for accurate relation recognition.
%
Furthermore, after being integrated with PR in Exp 3, our PE-Net obtains significant gains on mR@K (\eg, 33.8\% \vs 25.4\% at mR@100 on PredCls), which demonstrates PR's effectiveness in enlarging the distinction between prototypes achieving reliable entity-predicate matching.
However, we observe that the improvement of mR@K brings a slight drop at R@K. 
That is because our model can reasonably classify some predicates (head classes) into their corresponding fine-grained ones (tail classes), \ie, from ``on'' to ``standing on/laying on/walking on''. And the drops of Recall on those head predicates are inevitable, which is also observed in fine-grained classification~\cite{sgg:fgpl_a} and long-tailed tasks~\cite{lvis}.

\begin{figure}[t]
 \centering
 \includegraphics[width=1.0\linewidth]{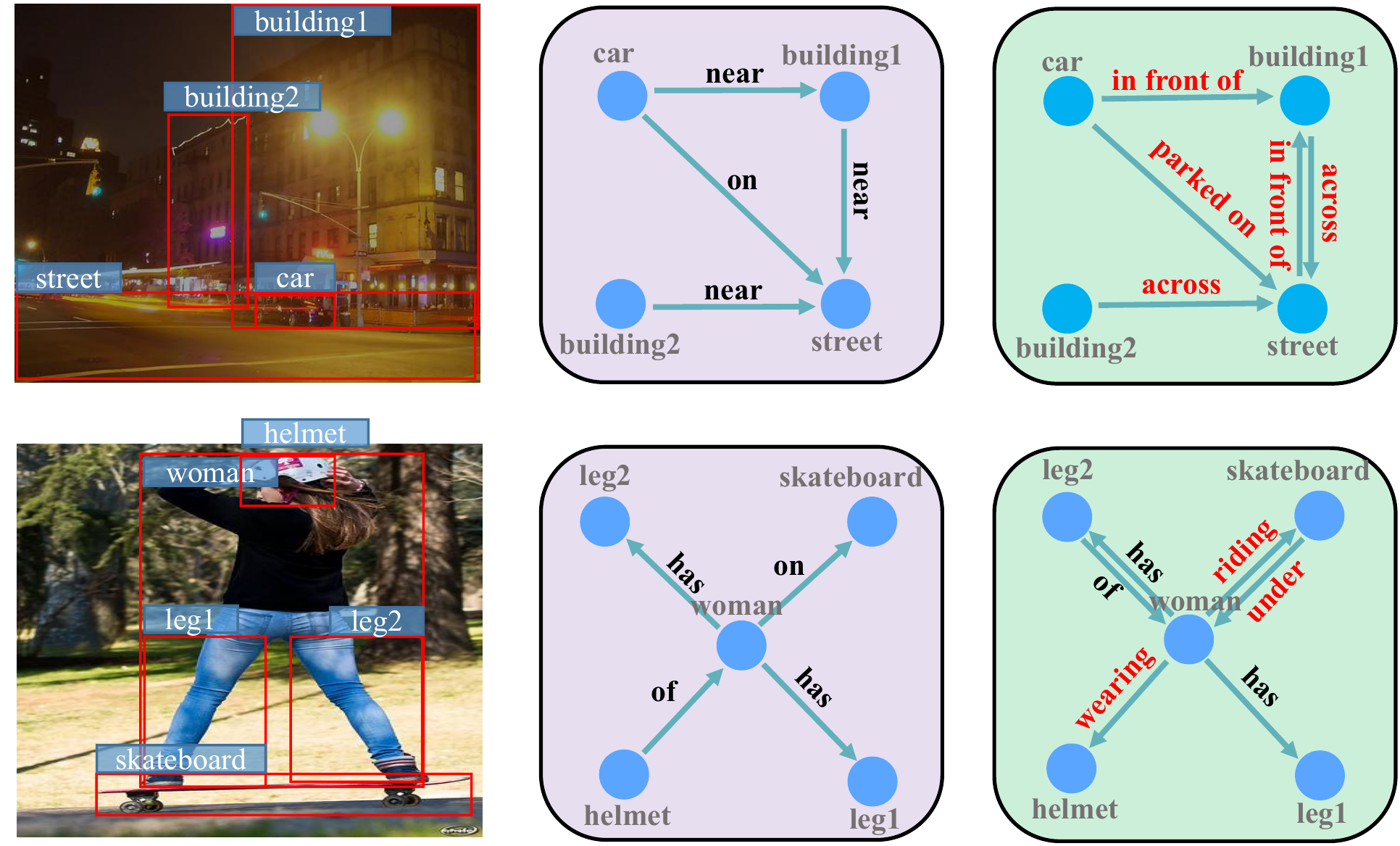} 
 \caption{Visualization Results of Motifs (in purple) and PE-Net (in green) on the PredCls task. 
 }
 \vspace{-1.2em}
 \label{fig:vis}
\end{figure}

\subsection{Visualization Results}
\label{subsec:visualization-res}
%
To verify that our proposed PE-Net is capable of making reliable relation recognition, we make a comparison between scene graphs generated by Motifs~\cite{sgg:motifs} (in purple) and our method (in green) in \cref{fig:vis}. 
In the first example, our method predicts informative relations such as ``car-parked on-street'' and ``building1-across-street'' instead of ``car-on-street'' and ``building1-near-street''.
Similarly, in the second example, our method generates fine-grained predicates, \eg, ``wearing'' and ``riding''.
These results demonstrate that our method has a stronger predicate recognition ability than Motifs,
which generates accurate relations for comprehensive scene understanding.

\section{Conclusion}  
In this work, we propose a novel Prototype-based Embedding Network (PE-Net), which produces compact and distinctive entity/predicate representations for SGG task. Towards this end, the PE-Net models entity and predicate instances with prototype-based representations and then matches entity pairs with predicates for relation recognition. Moreover, we propose a Prototype-guided Learning strategy (PL) and Prototype Regularization (PR) to help PE-Net efficiently learn entity-predicate matching. Finally, our method achieves new state-of-the-art performances on both Visual Genome and Open Images datasets, which demonstrates the effectiveness of our methods.

\noindent \textbf{Broader Impact and Limitations.}
Our work presents a powerful and efficient SGG method, which predicts relations between entities without message-passing module. The merit greatly reduces the computational complexity and enables SGG to be widely used in real-world applications, such as autonomous driving and intelligent robotics. However, our method is sensitive to the detector's recognition ability for entities, which limits its performance on SGCls and SGDet subtasks. Therefore, a more robust modeling method should be explored in further work.

\noindent \textbf{Acknowledgment.}
This study is supported by grants from National Key R\&D Program of China (2022YFC2009903/2022YFC2009900), the National Natural Science Foundation of China (Grant No. 62122018, No. 62020106008, No. 61772116, No. 61872064), Fok Ying-Tong Education Foundation(171106), SongShan Laboratory YYJC012022019, and Open Research Projects of Zhejiang Lab (No. 2019KD0AD01/011).

{\small
\bibliographystyle{ieee_fullname}
\bibliography{egbib}
}

\end{document}